\pdfoutput=1

\documentclass[11pt]{article}

\usepackage[]{acl}

\usepackage{times}
\usepackage{latexsym}

\usepackage[T1]{fontenc}

\usepackage[utf8]{inputenc}

\usepackage{microtype}

\usepackage{inconsolata}

\usepackage{makecell}
\usepackage{fontawesome}
\usepackage{makecell}

\usepackage{graphicx}
\usepackage{booktabs}
\usepackage{multirow}
\usepackage{multicol}
\usepackage{amssymb}

\usepackage{subfig}
\usepackage{geometry} 
\usepackage{marvosym}
\usepackage{microtype}
\usepackage{times}
\usepackage{latexsym}
\usepackage{tcolorbox}
\usepackage{booktabs}
\usepackage{siunitx}
\usepackage{colortbl}
\usepackage{color}
\usepackage{arydshln}
\usepackage{longtable}
\usepackage{array} 

\title{\M: Deepen and Broaden Large Language Model Assessment \\ via Structured Evaluation}

\author{Boxi Cao${}^{1,3}$, Mengjie Ren${}^{1,3}$, Hongyu Lin${}^{1}$, \textbf{Xianpei Han}${}^{1, 2,4}$\textsuperscript{\Letter}\\ \textbf{Feng Zhang}${}^{5}$, \textbf{Junfeng Zhan}${}^{5}$,\textbf{Le Sun}${}^{1,2,4}$\textsuperscript{\Letter}\\
${}^{1}$Chinese Information Processing Laboratory ~ ${}^{2}$State Key Laboratory of Computer Science \\
Institute of Software, Chinese Academy of Sciences, Beijing, China\\
${}^{3}$University of Chinese Academy of Sciences, Beijing, China \\
${}^{4}$Key Laboratory of System Software, Chinese Academy of Sciences \\
${}^{5}$ByteDance Inc. \\
{\tt \{boxi2020,hongyu,xianpei,sunle\}@iscas.ac.cn} 
}

\newcommand{\M}{StructEval}

\begin{document}
\maketitle
\begin{abstract}
Evaluation is the baton for the development of large language models (LLMs).  
Current evaluations typically employ a single-item assessment paradigm for each atomic test objective, which struggles to discern whether a model genuinely possesses the required capabilities or merely memorizes/guesses the answers to specific questions.
To this end, this paper proposes a novel evaluation framework referred to as \M. 
Starting from an atomic test objective, \M~deepens and broadens the evaluation by conducting a structured assessment across multiple cognitive levels and critical concepts, and therefore offers a comprehensive, robust and consistent evaluation for LLMs.
Experiments on three widely-used benchmarks demonstrate that \M~ serves as a reliable tool for resisting the risk of data contamination and reducing the interference of potential biases, thereby providing more reliable and consistent conclusions regarding model capabilities. 
Our framework also sheds light on the design of future principled and trustworthy LLM evaluation protocols\footnote{We openly release our source code and latest benchmark at \url{https://github.com/c-box/StructEval}, as well as leaderboard at \url{https://huggingface.co/spaces/Bowieee/StructEval_leaderboard}.}.
\end{abstract}

\section{Introduction}
\label{sec:intro}

Evaluation is fundamental for the development of large language models (LLMs)~\cite{ouyangTrainingLanguageModels2022a, touvronLlamaOpenFoundation2023a,openaiGPT4TechnicalReport2023}, providing essential measurements, feedback, and insights that facilitate enhancements in helpfulness, reliability and security~\cite{changSurveyEvaluationLarge2023}. 
Consequently, a variety of large-scale benchmarks are proposed to assess LLMs' capabilities, such as language understanding~\cite{hendrycksMeasuringMassiveMultitask2021, huangCEvalMultiLevelMultiDiscipline2023}, instruction following~\citep{alpaca_eval,zhengJudgingLLMasaJudgeMTBench2023a}, reasoning capabilities~\cite{cobbeTrainingVerifiersSolve2021, srivastavaImitationGameQuantifying2023}.

\begin{figure}[!tp]
\centering
    \includegraphics[width=\columnwidth]{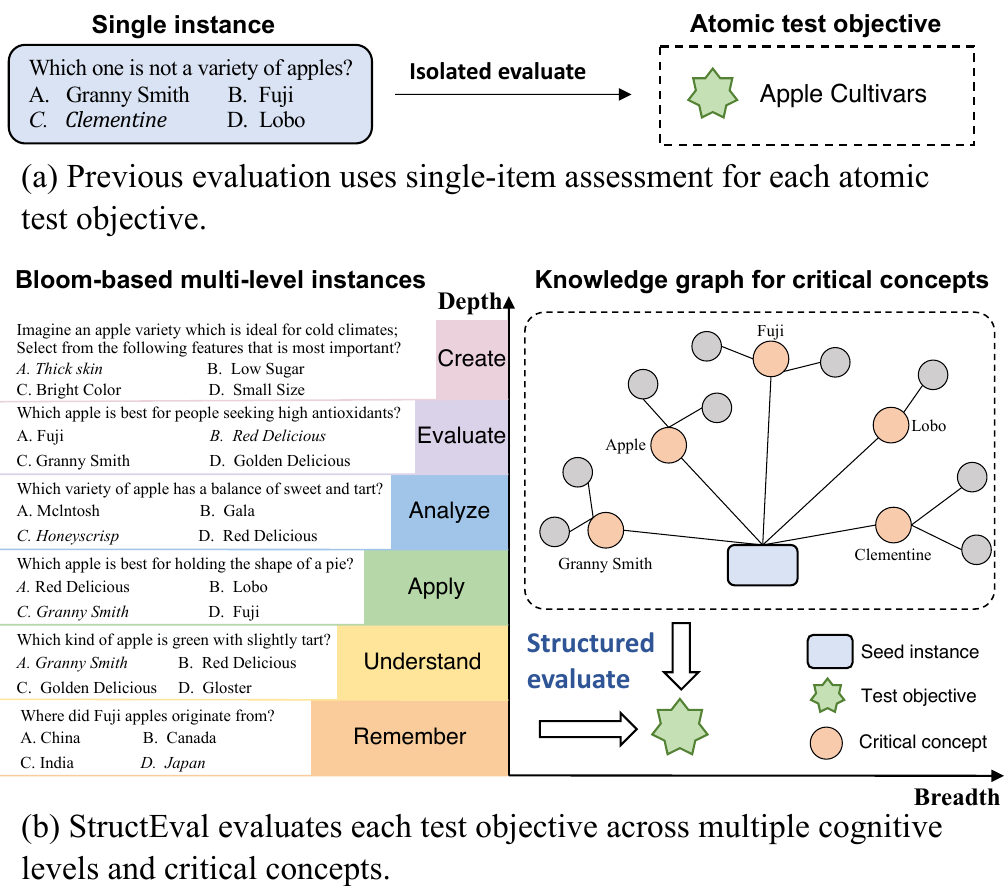}
    \caption{The illustrations for previous single-item assessment and our structured evaluation paradigm.}
    \label{fig:head}
\end{figure}

Unfortunately, current evaluations for LLMs typically employ a single-item assessment paradigm\cite{miltonReliabilityValidityTesting2011}, which still suffer from their weakness on validity, robustness and comprehensiveness.
As demonstrated in Figure~\ref{fig:head}a, to evaluate the factual knowledge in LLMs, they segment the factual knowledge into a set of atomic test objectives (e.g., \textit{apple cultivars, function of insulin}), and evaluate each with a single instance (e.g., \textit{which one is not a variety of apple}).
However, such a single-item assessment paradigm struggles to discern whether a model genuinely possesses the required capability or merely memorizes/guesses the answers to specific questions.
On the one hand, the single-item assessment relies on the correctness of isolated instances, which is sensitive to confounders correlated to specific instances~\cite{poernerEBERTEfficientYetEffectiveEntity2020a,zhuPromptBenchEvaluatingRobustness2023}, and susceptible to biases or shortcuts~\cite{caoCanPromptProbe2022, xieAskAgainThen2023,wangDecodingTrustComprehensiveAssessment2024}, making it difficult to discern whether a model's correct response is due to genuine understanding or mere memorization~\citep{caoKnowledgeableEducatedGuess2021a,cao-etal-2024-retentive-forgetful}.
On the other hand, the rapid expansion of LLMs' training data and memorization capacity have heightened the risk of data contamination in static benchmarks\cite{carliniQuantifyingMemorizationNeural2023,jiangInvestigatingDataContamination2024}, potentially leading to inflated evaluations of model capabilities~\cite{magarDataContaminationMemorization2022, orenProvingTestSet2023a, shiDetectingPretrainingData2023b}. 
That is, the true capabilities of the models might be overestimated owing to the potential contamination of the training dataset by test instances. 
Moreover, due to the huge resources required for benchmark construction, currently most benchmarks assess models in a static manner. Consequently, they may quickly reach saturation due to the inability to update in timeliness, complexity and diversity.

To address the aforementioned challenges, previous research has primarily attempted to manually construct newer, harder, and more diverse benchmarks. For instance, ~\citet{kasaiRealTimeQAWhat2022, yuKoLACarefullyBenchmarking2023} devised evaluation benchmarks drawing from recent news or articles;  
~\citet{wangAdversarialGLUEMultiTask2022,wangRobustnessChatGPTAdversarial2023} added perturbations into the original datasets to assess model robustness;
~\citet{hendrycksMeasuringMassiveMultitask2021, huangCEvalMultiLevelMultiDiscipline2023} collected test instances from human professional examination to increase difficulty and diversity.
Despite the substantial resource invested, the single-item assessment paradigm of previous benchmarks still struggles with determining whether the evaluated performance can faithfully and fairly reflect the capabilities of models.

In this paper, we propose a novel structured evaluation framework named \M, which can comprehensively, robustly and validly evaluate LLMs.
This is achieved by employing a structured assessment guided by pedagogy theories to evaluate model ability for each test objective across multiple cognitive levels and critical concepts, rather than relying on the correctness of a single test instance.
Specifically, as illustrated in Figure~\ref{fig:head}b, \M~ consists of two modules which deepen and broaden current evaluation respectively.
Given a seed instance, the first module identifies its underlying test objective, and then generates multiple test instances around this test objective which are aligned with the six cognitive levels outlined in Bloom’s Taxonomy~\cite{krathwohlRevisionBloomTaxonomy2002}. 
Meanwhile, the second module extracts the key concepts that must be understood to answer the seed question~\cite{trochimIntroductionConceptMapping1989}, and then develop a series of instances revolving around these concepts based on knowledge graph.
Unlike single-item assessment, for each test objective, \M~ requires LLMs to demonstrate knowledge across multiple cognitive levels and a thorough comprehension of critical concepts for good performance.
In this way, 
for each test objective, the assessment conclusion is no longer determined by the correctness of a single instance. 
As a result, it does not depend on confounders introduced by specific instances, such as prompt selection, surface form shortcut, data distribution, etc. 
Therefore, \M~ can reduce the impact of biases brought by these confounders, providing more consistent and accurate assessment conclusions for various LLMs.
Meanwhile, a model with data contamination can merely memorize specific answers but still lacks corresponding structured knowledge, therefore, \M~ can robustly provide stable assessment results even when the training data is contaminated. 
Moreover, due to \M's capability to automatically generate large-scale and high-quality instances, thereby realizing dynamic evaluation through updating of knowledge sources, it can also prevent benchmarks from rapidly reaching saturation. 

To demonstrate the effectiveness of our framework, we implement \M~ based on 3 widely used benchmarks. The experiments on a variety of LLMs demonstrate that \M: 
1) enables the automating generation of 
 large-scale benchmarks and completion of structured evaluations, while ensures instance correctness, relevance, and helpfulness.
2) effectively resists the risk of data contamination, providing robust evaluation results even under data contamination settings.
3) significantly enhances the consistency of model rankings across different experiments, offering more precise and stable conclusions from evaluations.
4) substantially outperforms previous augmentation-based strategies such as word perturbation, paraphrasing, back translation, option shuffle, etc.

\begin{figure*}[!tp]
\centering
    \includegraphics[width=0.95\textwidth]{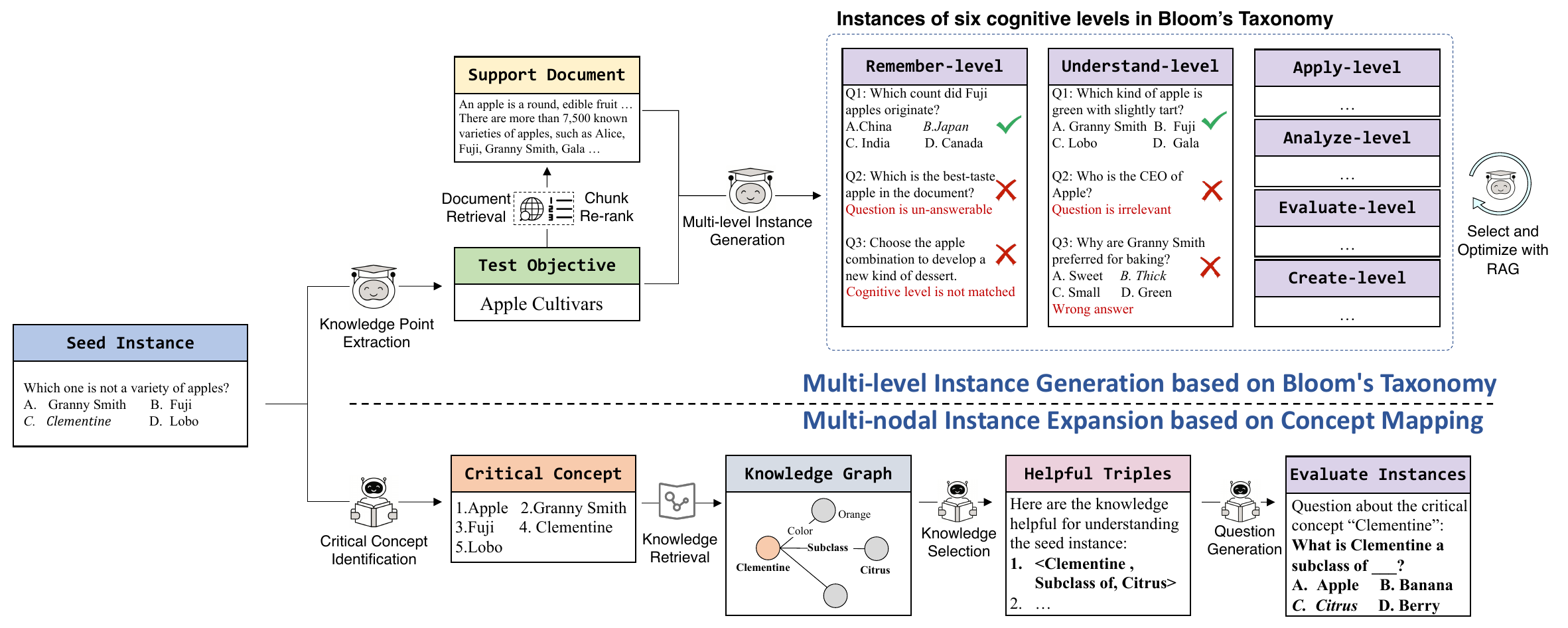}
    \caption{The illustration of \M~ framework, which consists of two modules. The first module aims to evaluate the model's ability on test objective across multiple cognitive levels in Bloom's Taxonomy. The second module aims to evaluate the model's understanding of relevant critical concepts based on knowledge graph.}
    \label{fig:frame}
\end{figure*}

The main contributions of this paper include:
\begin{itemize}
\item We propose a novel evaluation framework named \M, which can comprehensively evaluate LLMs' capability by assessing each test objective across multiple cognitive levels and critical concepts in principle, rather than previous single-item assessment.
\item We implement \M~ on widely-used benchmarks, and human evaluation results demonstrate that \M~ can automatically construct large-scale benchmark with high quality.
\item We conduct comprehensive experiments regarding data contamination and rank consistency, demonstrating the effectiveness, robustness and consistency of \M~ for LLM evaluation.
\end{itemize}

\section{Preliminaries}
\label{sec:pre}
Evaluation is the cornerstone for the progress of LLMs~\citep{changSurveyEvaluationLarge2023}.
Unfortunately, there still exist several grand challenges for achieving comprehensive and trustworthy evaluation for LLMs.
For instance, 
the inability to scale in complexity and diversity at the same pace as the rapid advancements in model capabilities\cite{srivastavaImitationGameQuantifying2023a,huangCEvalMultiLevelMultiDiscipline2023a};
the biases or shortcuts that lead to unfaithful assessments ~\cite{liangHolisticEvaluationLanguage2023,xieAskAgainThen2023};
and the lack of reliable metrics for providing trustworthy results~\cite{zhengJudgingLLMasaJudgeMTBench2023b,wangLargeLanguageModels2023}.
To this end, previous studies have mainly devoted to improving the diversity, scale, difficulty and timeliness of test instances~\citep{kasaiRealTimeQAWhat2022,zhuDyValGraphinformedDynamic2023}, exploring the robustness and trustworthiness vulnerabilities in current evaluations~\cite{zhuPromptBenchEvaluatingRobustness2023,wangDecodingTrustComprehensiveAssessment2024}, and proposing metrics or protocols more suitable for generative LMs~\citep{linLLMEvalUnifiedMultiDimensional2023,zhangWiderDeeperLLM2023a}. 
In comparison, this paper aims to propose a structured evaluation framework for LLM evaluation.

\M~ framework is guided by two pedagogy theories which are widely used for educational assessment.
\textit{Bloom's Taxonomy Theory} is a hierarchical model\footnote{We adopt the cognitive domain list since it is frequently used for educational assessment.} used for classification of educational learning objectives into six levels, including remember, understand, apply, analysis, evaluate and create~\cite{krathwohlRevisionBloomTaxonomy2002}. Therefore, to comprehensively evaluate the model's ability across various cognitive levels on test objective, \M~ would generate multiple test instances covering six cognitive levels in  Bloom's Taxonomy.
\textit{Concept Mapping theory} is another well-known tool for student assessment. Educators use concept maps to assess the breadth of a student's  understanding of a subject, which reveals how well students grasp connections among concepts~\cite{trochimIntroductionConceptMapping1989}. Therefore, to assess whether the model genuinely possess the knowledge required for test instance, \M~ would develop a series of instances revolving the critical concepts based on knowledge graph. 

\section{\M~ Framework}
\label{sec:framework}

The overall framework of \M~ is illustrated in Figure~\ref{fig:frame}, which consists of two modules. 
Given a seed instance, 
the first module would evaluate the model's ability on test objective across multi cognitive level. It first identifies the underlying test objective of this instance and then generates multiple relevant instances covering six cognitive levels of Bloom's Taxonomy.
The second module evaluates the model's comprehensive understanding of all critical concepts related to the seed instance. 
It extracts the essential concepts that must be understood and develops a series of extended questions around these concepts using a knowledge graph.
In the following, we would describe the \M~ framework in detail.

\subsection{Bloom's Taxonomy-based Instance Generation}

As shown in Figure~\ref{fig:frame}, given a seed instance, the first module of \M~automatically generates test instances corresponding to the six cognitive levels in Bloom's Taxonomy through the following steps: 1) extract the test objective examined by the seed instance; 2) retrieve relevant documents and re-rank the document chunks based on their relevance to seed instance; 3) generate candidate evaluation instances for each cognitive level in Bloom's Taxonomy using in-context learning; 4) select instances that best meet the requirements and refine them to be more challenging.
Subsequently, each component will be introduced in detail.

\paragraph{Test Objective Extraction} aims to identify the underlying test objective for each seed instance. 
For example, the test objective for question ``\textit{Which one is not a variety of apple?}'' is ``\textit{apple cultivars}''. 
However, such a single question is insufficient to thoroughly evaluate the LLM's related knowledge.
Therefore, to comprehensively assess how much knowledge the LLM possesses about the test objective and its level of understanding across different cognitive tiers, we conduct a structured evaluation around this test objective.
In our framework, we prompt LLM with few-shot demonstration to extract the test objective examined by each instance in the benchmarks.

\paragraph{Relevant Document Retrieval} 
Given the test objective corresponding to the seed instance, an intuitive approach is directly prompting LLM to generate instances for each cognitive level. 
However, this approach is severely compromised by the LLM's hallucinations, resulting in a significant proportion of incorrect instances. 
Therefore, \M~would first retrieve relevant passages, and then re-rank document chunks based on the correlation with the seed instance. 
This procedure ensures that the generation of subsequent instances is firmly based on the retrieved context, guaranteeing the precision and pertinence of the generated instances.

\begin{table}[t]
    \centering
    \small
    \begin{tcolorbox}
    \textbf{\# Background} \\
    Bloom's Taxonomy categorizes educational objectives into six levels. Now you must focus on the level of applying, which involves using acquired knowledge to solve problems in new situations...
    \\

    \textbf{\# Instruction} \\
    For a given test objective and document, generate 5 instances based on the following principles: \\
    1. Ensure the question can be answered independently without additional context. \\
    2. Ensure the correct answer and supporting evidence are available within the provided document. \\
    3. Ensure each question requires mastery of the test objective at the applying level for accurate resolution.
    \\
    
    \textcolor{blue}{\textbf{Few shot demonstrations}}
    \\
    
    \textbf{\# User Input} 
    \\
    test objective: <objective>\\
    Relevant Documents: <documents>\\
    5 instances of applying level in Bloom's Taxonomy:

    \end{tcolorbox}
    \caption{Prompt design for candidate instances generation. Please refer to the Appendix~\ref{app:detail} for more framework design details.}
    \label{tab:prompt_bloom}
\end{table}

\paragraph{Candidate Instances Generation} aims to generate multiple candidate instances for each cognitive level in Bloom's Taxonomy, based on the  test objective with relevant document chunks.
As demonstrated in Table~\ref{tab:prompt_bloom}, we meticulously design the prompt for LLM to generate relevant, correct and helpful instances corresponding to each cognitive level.
The prompt begins with introducing Bloom's Taxonomy and current cognitive level, following by the task instruction which includes three principles to ensure the answerability, accuracy and relevance of the generated instances.
Subsequently, we provide manually created few-shot demonstrations, and ask LLM to generate candidate instances using these demonstrations as references.

\paragraph{Instance Selection and Refinement}
Since the quality and difficulty of these instances may vary greatly, as shown in Figure~\ref{fig:frame}, we introduce a post-processing module aimed at selecting the highest quality instances for each cognitive level. 1) To ensure the answerability and correctness of instances, we prompt LLM to eliminate questions necessitating specific contextual information for resolution, and employ Retrieval-Augmented Generation (RAG) module to exclude questions that cannot be correctly answered based on the provided context, thereby ensuring the accuracy of the generated answers; 2) To enhance the quality and difficulty of instances, inspired by ~\citet{clarkThinkYouHave2018a,linTruthfulQAMeasuringHow2022a}, we establish a comprehensive pool of diverse LMs. Questions that all models could answer correctly were eliminated, thus ensuring the discriminative efficacy.

Ultimately, for each instance within the original benchmark, we develop a hierarchical evaluation system capable of extensively assessing the tested model's knowledge across all six cognitive levels in Bloom's Taxonomy.

\subsection{Concept Mapping-based Instance Expansion}
The second module evaluates LLMs' knowledge for each test objective with a concept map.
The hypothesis behind is also intuitive: if a model genuinely possesses the necessary knowledge to answer a given question, it should demonstrate a comprehensive understanding of the critical relevant concepts. 
Specifically, as illustrated in Figure~\ref{fig:frame}, \M~ utilizes LLM and knowledge graph to expand the breadth of existing benchmarks with following steps:
1) Identify the key concepts that must be understood to correctly answer the seed question;
2) Retrieve relevant knowledge sub-graphs for each concept and select the necessary knowledge triplets from all the candidates for understanding the original question;
3) Transform the selected triplets into test instances and optimize their difficulty.

\paragraph{Critical Concept Identification} aims to identify the critical concepts that must be understood to correctly answer the seed question.
These concepts are then linked to the entries in knowledge graph to facilitate subsequent knowledge retrieval.
Previous approaches such as BLINK~\cite{wuScalableZeroshotEntity2020} are constrained on the entity label set and fail to discern between critical and non-critical concepts.
Therefore, we prompt LLM with few-shot demonstration to identify the critical concepts in instances.

\paragraph{Knowledge Graph Retrieval and Selection}
 involves retrieving the identified critical concepts across the entire knowledge graph and extracting relevant knowledge triples from the sub-graph as candidates.
Given the potential enormity of the candidate set, which may contain extraneous triplets not aiding in determining the model's ability to answer the seed question, similar to~\citet{guanMitigatingLargeLanguage2023a}, we prompt the LLM to select the helpful knowledge triplets with few-shot demonstrations.

\paragraph{Instances Generation and Optimization} transforms the selected factual triples into evaluation instances. Similar to~\citet{petroniLanguageModelsKnowledge2019a}, we utilize the subject entity and its relation to formulate the question, with the object entity as the answer.
For multiple-choice questions, in order to ensure the difficulty of the questions, we first use the taxonomy of the knowledge graph to identify the finest-grained entity category corresponding to the correct answer. Then, we select the incorrect options from other entities within the same category.

Ultimately, we construct a multi-nodal evaluation framework for each test instance, offering a comprehensive assessment of the language model’s grasp of pertinent critical concepts.

\section{Implementations and Experiments of \M~}
\label{sec:experiments}

In this section, 
we first implement \M~ across three widely-used benchmarks. Through human evaluation, we demonstrate capability of \M~ to automatically construct large-scale benchmarks while ensuring the helpfulness, answerability and correctness of generated instances.
Then, we demonstrate how \M~ could improve the robustness and consistency of LLM evaluation from the following perspectives.
Firstly, \M~ requires LLMs to understand the test objective across multiple cognitive levels and critical concepts. In this case, a contaminated model which merely memorize specific answers may achieve high performance in original benchmark, but cannot gain performance improvements on the structured evaluation since it lacks of corresponding knowledge. 
Therefore, \M~ can effectively resist data contamination issues, providing robust evaluation results even when the test data is leaked. 
Secondly, since the evaluation results do not rely on the correctness on single instance, it does not depend on confounders introduced by specific instances, such as prompt selection, surface form shortcut and data distribution.
Therefore, compared with single-item assessment, \M~ can provide a much more robust and consistent evaluation conclusion.

\subsection{\M~-based Benchmarks}
\label{sec:bench}

\textbf{Finding 1.} \emph{By leveraging the advanced generative capabilities of LLMs, and meticulously orchestrating the construction process guided by pedagogy theories and grounded in credible knowledge sources, \M~ is capable of automatically construct large-scale benchmarks while ensuring the helpfulness, answerability and accuracy of generated instances.}

To demonstrate the reliability and quality of the automatically generated instances by \M, we adopt \M~ on three widely used benchmarks for LLMs, and conduct human evaluations from three aspects.

\paragraph{Seed Benchmarks} include the following three wildly used LLM evaluation benchmarks, and the corresponding statistics are demonstrated in Table~\ref{tab:bench_stat}: 1) \emph{MMLU}~\cite{hendrycksMeasuringMassiveMultitask2021} is a large-scale benchmark designed to measure the knowledge in large language models. We filtered out subjects unsuitable for ~\M, such as logical reasoning and numerical computation, and used the remaining 48 subjects for experiments. 
2) \emph{ARC}~\cite{clarkThinkYouHave2018a} consists of a set of science exam questions drawn from a variety of sources, which is widely used to assess the knowledge in LLMs. The benchmark is partitioned into a challenge set and easy set, and we include both of them in our experiments. 3) \emph{OpenBook QA}~\cite{mihaylovCanSuitArmor2018} is a question-answering datasets that consists of multiple-choice elementary-level science questions. 
For each benchmark, we randomly sample 200 test instances generated by \M~ to conduct human evaluations.

\paragraph{LLM and Knowledge Source Implementation} Considering the balance between cost, efficiency, and the quality of generation, we select ChatGPT\footnote{Gpt-3.5-turbo-1106 is used in the experiments of this paper, and GPT-4o-mini is used for our latest benchmark construction.}  for LLM generation tasks in this paper, and our framework can also be easily adopted to other large models and knowledge sources. 
We adopt BM25~\cite{maronRelevanceProbabilisticIndexing1960} for document retrieval and BGE~\cite{xiaoCPackPackagedResources2023} for chunk re-ranking.
We select Wikipedia~\citep{wikipedia} as knowledge source since it encompasses the vast majority information about test objectives and possesses an high density of knowledge, and use Wikidata~\citep{VrandecicKroetzsch14cacm} for fact retrieval, since it is one of the most comprehensive knowledge bases covering structured knowledge.

\paragraph{Metrics} 
We propose the following metrics to comprehensively assess the quality of test instances generated through ~\M:
1) \emph{Instance Helpfulness}, which is calculated by the proportion of generated evaluation instances that conform to the target test objective or critical related concepts; 2) \emph{Question Answerability}, which is calculated by the proportion of generated questions that can be answered without relying on external context. 3) \emph{Answer Correctness}, which is calculated by the proportion of generated evaluation instances that contain the correct answer.
The detailed annotation guideline is presented in the appendix.

\begin{table}[tp]
    \centering
    \resizebox{\columnwidth}{!}{
    \begin{tabular}{lcccc}
    \toprule
    \textbf{Benchmark}      & \textbf{Original} & \textbf{Bloom} & \textbf{Concept} & \textbf{Structured} \\ \hline
    MMLU           & 13.1k             & 135.8k         & 33.1k            & 168.9k              \\
    ARC-Easy       & 2.3k              & 17.8k          & 6.2k             & 24.0k               \\
    ARC-Challenge & 1.1k              & 14.2k          & 4.2k             & 18.4k               \\
    OpenBook QA    & 0.5k              & 6.0k           & 1.2k             & 7.2k               \\
    \bottomrule

    \end{tabular}}
    \caption{Data statistics of the seed benchmarks and \M~-constructed benchmarks. ``Bloom'' denotes the multi-level instances based on Bloom's taxonomy, ``Concept'' denotes the test instances about critical concepts, ``Structured'' indicates the comprehensive structured assessment that includes both components.}
    \label{tab:bench_stat}
\end{table}

\begin{table}[tp]
\centering
\resizebox{\columnwidth}{!}{
\begin{tabular}{lccc}
\toprule
\textbf{Benchmark} & \textbf{Helpfulness} & \textbf{Answerability} & \textbf{Correctness} \\ \hline
Struct MMLU     & 95.5  & 96.0  & 97.0  \\ 
Struct ARC    & 96.0 & 96.0 & 98.0\\
Struct OpenBook    & 97.5   & 95.0  & 94.5 \\
\bottomrule
\end{tabular}
}
\caption{The human evaluation results three benchmarks constructed by \M.}
\label{tab:quality}
\end{table}

\paragraph{Benchmark Assessment}
From table~\ref{tab:bench_stat}, we find that \M~ is able to automatically construct a large-scale multi-level and multi-nodal evaluation system based on original benchmark, provide novel test instances and 
 structured evaluation protocol for existing benchmarks.
Moreover, the human evaluation results are shown in table~\ref{tab:quality}. We can find that \M~ could construct the structured evaluations  while ensure the high quality of generated instances in all the aspects of instance helpfulness, question answerability, and answer correctness.
The few errors mainly due to that GPT-3.5 generate questions relying on context to answer or with multiple correct choices,
we also provide a detailed error analysis and the annotation guideline in the Appendix~\ref{app:error} due to page limitations.

\begin{table*}[!tp]
\centering
\resizebox{\textwidth}{!}{
\begin{tabular}{lccccccccccccccc}
\toprule
 &
  \multicolumn{3}{c}{\textbf{LLaMa-7B}} &
  \multicolumn{3}{c}{\textbf{LLaMa-30B}} &
  \multicolumn{3}{c}{\textbf{LLaMa-2-7B}} &
  \multicolumn{3}{c}{\textbf{LLaMa-2-13B}} &
  \multicolumn{3}{c}{\textbf{Mistral-7B}} \\ \cmidrule(r){2-4} \cmidrule(r){5-7} \cmidrule(r){8-10} \cmidrule(r){11-13} \cmidrule(r){14-16}
\multirow{-2}{*}{\textbf{Method}} &
  Clean IFT &
  w/ Test $\Downarrow$ &
  $\Delta$ $\Downarrow$ &
  Clean IFT &
  w/ Test $\Downarrow$ &
  $\Delta$ $\Downarrow$ &
  Clean IFT &
  w/ Test $\Downarrow$ &
  $\Delta$ $\Downarrow$ &
  Clean IFT &
  w/ Test $\Downarrow$ &
  $\Delta$ $\Downarrow$ &
  Clean IFT &
  w/ Test $\Downarrow$ &
  $\Delta$ $\Downarrow$  \\ \hline 
\multicolumn{16}{c}{\cellcolor[HTML]{EFEFEF}\textbf{MMLU}}                                                                               \\ 
Original          & 50.22 & 79.32 & +29.10 & 59.06 & 92.22 & +33.16 & 54.42 & 86.88 & +32.46 & 59.27 & 90.98 & +31.71 & 55.84 & 95.67 & +39.83 \\
CharDisturb       & 49.52 & 76.90 & +27.38 & 58.38 & 90.76 & +32.38 & 53.41 & 83.99 & +30.58 & 57.88 & 88.86 & +30.98 & 55.28 & 94.04 & +38.76 \\
Wordnet           & 49.07 & 75.76 & +26.69 & 57.80 & 90.11 & +32.31 & 53.58 & 83.25 & +29.67 & 58.06 & 88.16 & +30.10 & 54.95 & 93.38 & +38.43 \\
Paraphrasing      & 50.19 & 73.09 & +22.84 & 58.74 & 89.33 & +30.59 & 54.50 & 80.76 & +26.26 & 58.20 & 86.05 & +27.85 & 55.48 & 92.79 & +37.31 \\
BackTranslation   & 50.36 & 76.86 & +26.50 & 58.71 & 90.95 & +32.24 & 54.22 & 84.21 & +29.99 & 59.23 & 89.27 & +30.04 & 55.95 & 94.53 & +38.58 \\
OptionShuffle     & 50.70 & 66.50 & +15.80 & 58.69 & 83.69 & +25.00 & 53.84 & 72.57 & +18.73 & 58.94 & 78.04 & +19.10 & 54.97 & 84.21 & +29.24 \\
StructEval (\textit{ours}) &
  47.60 &
  49.33 &
  \textbf{+1.73} &
  56.87 &
  57.09 &
  \textbf{+0.22} &
  51.42 &
  53.24 &
  \textbf{+1.82} &
  56.53 &
  57.32 &
  \textbf{+0.79} &
  52.28 &
  53.95 &
  \textbf{+1.67} \\ \hline
\multicolumn{16}{c}{\cellcolor[HTML]{EFEFEF}\textbf{ARC-challenge}}                                                                       \\
Original          & 53.86 & 91.77 & +37.91 & 69.55 & 98.54 & +28.99 & 61.75 & 96.83 & +35.08 & 72.04 & 98.97 & +26.93 & 66.72 & 99.83 & +33.11 \\
CharDisturb       & 52.47 & 88.99 & +36.52 & 68.26 & 97.27 & +29.01 & 58.19 & 94.71 & +36.52 & 69.88 & 97.27 & +27.39 & 64.68 & 99.15 & +34.47 \\
Wordnet           & 50.51 & 86.18 & +35.67 & 65.36 & 95.05 & +29.69 & 58.36 & 92.75 & +34.39 & 67.75 & 96.33 & +28.58 & 61.95 & 97.27 & +35.32 \\
Paraphrasing      & 52.56 & 82.00 & +29.44 & 68.69 & 95.39 & +26.70 & 59.04 & 89.51 & +30.47 & 70.14 & 93.34 & +23.20 & 64.51 & 96.76 & +32.25 \\
BackTranslation   & 53.84 & 88.05 & +34.21 & 66.98 & 97.27 & +30.29 & 60.84 & 94.97 & +34.13 & 69.80 & 97.44 & +27.64 & 64.85 & 98.55 & +33.70 \\
OptionShuffle     & 53.84 & 76.37 & +22.53 & 70.22 & 93.26 & +23.04 & 62.12 & 83.36 & +21.24 & 70.05 & 90.02 & +19.97 & 65.44 & 94.45 & +29.01 \\
StructEval (\textit{ours}) &
  44.90 &
  44.71 &
  \textbf{-0.19} &
  54.49 &
  55.23 &
  \textbf{+0.74} &
  48.96 &
  49.69 &
  \textbf{+0.73} &
  54.58 &
  55.40 &
  \textbf{+0.82} &
  52.56 &
  52.27 &
  \textbf{-0.29} \\ \hline
\multicolumn{16}{c}{\cellcolor[HTML]{EFEFEF}\textbf{ARC-easy}}                                                                            \\
Original          & 77.06 & 97.22 & +20.16 & 86.95 & 99.49 & +12.54 & 79.38 & 99.16 & +19.78 & 85.40 & 99.62 & +14.22 & 81.73 & 99.96 & +18.23 \\
CharDisturb       & 74.24 & 94.70 & +20.46 & 85.40 & 98.19 & +12.79 & 76.52 & 97.31 & +20.79 & 83.21 & 97.94 & +14.73 & 79.17 & 99.28 & +20.11 \\
Wordnet           & 73.40 & 93.60 & +20.20 & 83.84 & 97.10 & +13.26 & 75.08 & 95.58 & +20.50 & 81.52 & 96.93 & +15.41 & 78.37 & 98.06 & +19.69 \\
Paraphrasing      & 75.21 & 93.01 & +17.80 & 86.24 & 97.43 & +11.19 & 78.24 & 95.33 & +17.09 & 83.33 & 96.80 & +13.47 & 81.69 & 98.65 & +16.96 \\
BackTranslation   & 75.84 & 95.12 & +19.28 & 84.72 & 98.06 & +13.34 & 77.27 & 97.31 & +20.04 & 83.16 & 98.19 & +15.03 & 80.26 & 99.12 & +18.86 \\
OptionShuffle     & 75.72 & 90.74 & +15.02 & 86.20 & 96.76 & +10.56 & 78.41 & 93.39 & +14.98 & 85.27 & 96.21 & +10.94 & 81.52 & 98.15 & +16.63 \\
StructEval (\textit{ours}) & 45.21 & 45.37 & \textbf{+0.16}  & 54.14 & 55.34 & \textbf{+1.20}  & 48.60 & 49.60 & \textbf{+1.00}  & 54.25 & 54.95 & +0.70  & 51.23 & 51.79 & \textbf{+0.56}  \\ \hline
\multicolumn{16}{c}{\cellcolor[HTML]{EFEFEF}\textbf{OpenBook QA}}                                                                         \\
Original          & 55.86 & 87.03 & +31.17 & 56.11 & 97.51 & +41.40 & 62.09 & 95.76 & +33.67 & 66.83 & 99.50 & +32.67 & 69.58 & 99.00 & +29.42 \\
CharDisturb       & 48.40 & 82.20 & +33.80 & 51.20 & 95.00 & +43.80 & 53.40 & 89.20 & +35.80 & 60.80 & 94.60 & +33.80 & 61.80 & 97.20 & +35.40 \\
Wordnet           & 49.00 & 79.40 & +30.40 & 52.20 & 92.60 & +40.40 & 55.20 & 86.80 & +31.60 & 57.40 & 93.40 & +36.00 & 61.00 & 96.60 & +35.60 \\
Paraphrasing      & 59.00 & 77.20 & +18.20 & 57.20 & 92.80 & +35.60 & 59.80 & 84.80 & +25.00 & 67.60 & 93.20 & +25.60 & 68.20 & 95.60 & +27.40 \\
BackTranslation   & 51.80 & 84.80 & +33.00 & 52.00 & 95.00 & +43.00 & 58.80 & 90.40 & +31.60 & 62.40 & 96.00 & +33.60 & 65.60 & 97.00 & +31.40 \\
OptionShuffle     & 53.60 & 87.60 & +34.00 & 53.80 & 97.60 & +43.80 & 59.40 & 95.40 & +36.00 & 64.60 & 99.60 & +35.00 & 67.60 & 98.80 & +31.20 \\
StructEval (\textit{ours}) &
  44.31 &
  42.57 &
  \textbf{-1.74} &
  51.97 &
  51.71 &
  \textbf{-0.26} &
  48.86 &
  46.89 &
  \textbf{-1.97} &
  55.48 &
  55.87 &
  \textbf{+0.39} &
  51.00 &
  49.64 &
  \textbf{-1.36} \\ \bottomrule
\end{tabular}
}
\caption{Performance comparisons of LLMs which are trained on clean data and contaminated data. ``w/Test'' indicates that the instruction tuning data is contaminated by the test samples. ``$\Delta$'' denotes the performance divergence between clean and contamination settings, lower values ($\Downarrow$) reflect that the corresponding evaluation is less affected by data contamination. The evaluation is conducted under zero-shot setting.}
\label{tab:data_con_res}
\end{table*}

\subsection{Robustness of \M}
\label{ssec:data_con}

\textbf{Finding 2.} \emph{By expanding the benchmark across both depth and breadth dimensions, \M~ is capable of robustly evaluating the capabilities of LLMs, resisting the risks of data contamination, and providing stable results even under data contamination settings.}

Data contamination refers to the inclusion of test data in the training dataset of evaluated models, which can significantly skew the apparent performance and capabilities of models, leading to misleading conclusions about their true effectiveness.
Addressing data contamination becomes increasingly crucial for large language models as the training data grow exponentially with the data sources and processing recipes being obscure.

To demonstrate the effectiveness of \M~ in resisting the risk of data contamination, 
we compare the performance divergences of LLMs with and without data contamination, on the original benchmark, the data augmented benchmark and \M-constructed benchmark respectively. 
Specifically, for a seed benchmark and a base model, we use instruction fine-tuning (IFT) to train the model on both a clean dataset and a dataset contaminated with test data.
To make a fair comparison, we ensure that both datasets maintain identical scale and similar data composition. 
Simultaneously, we integrate Alpaca-GPT-4~\citep{alpaca} dataset into both the training data to ensure data diversity and prevent training collapse.
In this case, the \textbf{contaminated set} consists of Alpaca-GPT-4 and the test data, while the \textbf{clean set} consists of Alpaca-GPT-4 and an equal number of multi-choice questions which are randomly sampled from an out-of-distribution benchmark Xiezhi~\cite{guXiezhiEverUpdatingBenchmark2023}.
To ensure the robustness of our conclusions, we consider 5 wildly used base LLMs of various scales including LLaMa-7B\&30B\cite{touvronLLaMAOpenEfficient2023}, LLaMa-2-7B\&13B\cite{touvronLlamaOpenFoundation2023a} and Mistral-7B\cite{jiangMistral7B2023}.
Each model is trained through 3 epochs with batch size of 256 sequences, using Adam with learning rate $2e-5$.

We also compare our method with the following augmentation-based approaches including character-level, word-level and instance-level, which are able to generate adversarial samples while ensuring the answerability and correctness of test instances:
1) CharDisturb~\citep{morris2020textattack}: which randomly substitutes, deletes, inserts and swaps characters in original question.
2) WordNet~\cite{millerWordNetLexicalDatabase1995a}, which randomly replaces words with WordNet synonyms.
3) Paraphrasing~\cite{zhuCLEANEVALCleanEvaluation2023}, which prompts ChatGPT to generate paraphrasing for each test question. 
4) BackTranslation~\cite{sennrichImprovingNeuralMachine2016}, which translates the test question into another language and translates it back.
5) OptionShuffle~\cite{yangRethinkingBenchmarkContamination2023}, which re-ordered the options for each question to prevent LLMs memorizing specific option for question.

The results in Table~\ref{tab:data_con_res} clealy demonstrate the significant role of \M~ in resisting data contamination:
1) \textbf{The performance of original benchmark can severely suffer from data contamination} due to the superior memorizing capabilities of LLMs, resulting in a serious overestimation of the model's capabilities. For instance, the performance of all models on MMLU increase by at least $29\%$ when the training data is contaminated
2) \textbf{Previous augmentation-based approaches struggle to resist data contamination.} 
Despite adjustments to the surface form of the original instances, due to the LLMs' advanced memorizing and language comprehension capabilities, they still achieve significant benefits from data contamination.
3) \textbf{\M~ is able to provide stable evaluation results, regardless of whether the training data is contaminated.}
For example, due to data contamination, the performance of LLaMa-2-13B improves by $31.71\%$ on the original MMLU, but changes by only $0.79\%$ on the structured-MMLU generated by \M, which remains almost unchanged.
The finding remains consistent across all base LLMs and benchmarks.
Such results effectively demonstrate that \M~ can play a role in anti-attack and contamination monitoring for evaluation.

\subsection{Consistency of \M}
\label{ssec:consis}

\begin{table}[tp]
\centering
\resizebox{\columnwidth}{!}{
\begin{tabular}{lcccccc}
\hline
\textbf{Model} & \textbf{Ori} & \textbf{Char} & \textbf{Word} & \textbf{Trans} & \textbf{Shuffle} & \textbf{Struct} \\ \hline
Mistral-7B$^\star$        & 59.74  & 57.09  & 54.32  & 45.35  & 42.08  & \textbf{61.86}  \\
Mistral-8*7B$^\dag$   & 46.73  & 50.38  & 52.16  & 55.40  & 66.05  & \textbf{100.00} \\
LLaMA-7B$^\star$          & 99.91  & 99.95  & 99.14  & 99.89  & 95.58  & \textbf{99.99}  \\
LLaMA-30B$^\star$          & 38.97  & 44.84  & 43.42  & 41.58  & 41.90  & \textbf{70.32} \\
LLaMA-2-7B$^\star$         & 37.31  & 40.02  & 41.77  & \textbf{59.31}  & 36.07  & 48.39  \\
LLaMA-2-13B$^\star$        & 38.29  & 49.63  & 44.20  & 46.64  & 48.89  & \textbf{76.80}  \\
Baichuan2-7B$^\dag$   & 45.37  & 49.93  & 85.71  & 67.10  & 57.89  & \textbf{94.42}  \\
Baichuan2-13B$^\dag$  & 55.32  & 41.78  & 58.00  & \textbf{64.70}  & 40.98  & 53.93  \\
Qwen-7B$^\dag$       & 40.09  & 40.70  & 48.65  & 47.32  & 32.59  & \textbf{94.41}  \\
Qwen-14B$^\dag$       & 87.18  & \textbf{88.27}  & 84.33  & 84.88  & 71.44  & 86.86  \\
Qwen1.5-7B$^\dag$     & 51.55  & 40.46  & 42.84  & 49.03  & 34.27  & \textbf{98.59}  \\
Qwen1.5-14B$^\dag$    & 57.84  & 59.95  & 52.54  & 55.35  & 64.26  & \textbf{100.00}  \\
Yi-6B$^\dag$          & 59.76  & 68.68  & 60.49  & 40.91  & 52.05  & \textbf{100.00}  \\ \hline
Overall Rank        & 1.24   & 1.63   & 3.15   & 2.28   & 1.48  & \textbf{33.17} \\ \hline
\end{tabular}
}
\caption{The \textit{rank consistency} of each LLM over 10000 task samples, and each task consists of $K=15$ subjects from MMLU. ``Overall Rank'' indicates percentage of the most popular rank of all models across 10000 runtimes. ``$\star$'' denote that the base model is trained with clean IFT. ``$\dag$'' denote the chat version of model. }
\label{tab:rank_con}
\end{table}

\textbf{Finding 3.} \emph{By conducting structured assessments across various cognitive levels and essential concepts, instead of basing assessments solely on the accuracy of a single instance, \M~ achieves 
 valid assessment of models, providing consistent conclusions regarding various model capabilities.}

As we discussed above,
\M~ can also serve as a more stable reference for assessing the knowledge capabilities of language models, which can give more stable evaluation results to various LLMs, and reach a consistent conclusion.
Demonstrating this requires to collect numerous benchmarks with similar evaluation objective and distribution, and observe whether the evaluation conclusions are consistent on original data, augmented data and \M-constructed data respectively. 
To facilitate our experiments, we refer to ~\citet{caoCanPromptProbe2022} and use \emph{rank consistency} across multiple runtimes as the evaluation metric. 
Specifically, we randomly sample 10000 sub-set with $K$ subjects from MMLU, and evaluate rank consistency by measuring the percentage of the most popular rank of each model in 10000 runtimes.
For instance, if ChatGPT ranks at $3^{rd}$ place in 6500 of the 10000 runtimes, then the rank consistency of ChatGPT would be $65\%$.
To make a comprehensive evaluation, we conduct experiments on 13 different open-source large language models across various parameter scales, including LLaMA-7B\&30B, LLaMA-2-7B\&13B, Mistral-7B\&8*7B, Baichuan2-7B\&13B\cite{baichuan2023baichuan2}, Qwen-7B\&14B, Qwen1.5-7B\&14B\cite{qwen} and Yi-6B.
We report the rank consistency of each model, as well as the rank consistency across all models.

\begin{figure}[!tp]
\centering
    \includegraphics[width=0.95\columnwidth]{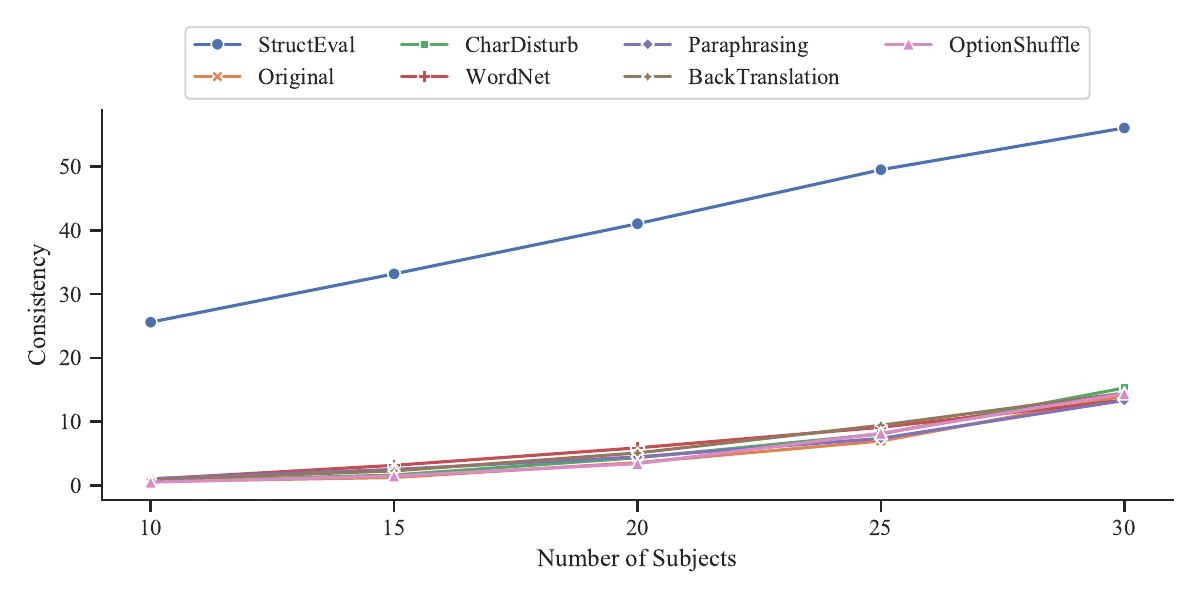}
    \caption{The comparison of \textit{overall rank consistency} for each method. \M~ substantially outperforms original benchmark and all augmentation-based strategies as number of sampled subjects $K$ changes.}
    \label{fig:consis_line}
\end{figure}

The results in Table~\ref{tab:rank_con} and Figure~\ref{fig:consis_line} demonstrate that \M~ can significantly improve the evaluation consistency: 1) \textbf{The consistency of current LLM evaluations are relatively poor}: when using original isolated instances to compare the ability of different models, the overall rank consistency is only $1.24\%$. 
2) \textbf{Previous strategies can hardly improve the rank consistency.}
Although they modify the original data, they still adhere to the paradigm of single-item assessment. As a result, they remain susceptible to interference from confounders and struggle to provide more consistent evaluation conclusions across all models.
3) \textbf{\M~ provides much more consistent evaluation conclusions regarding the ability of different LLMs}: the overall rank consistency improved from $1.24\%$ to $33.17\%$ when $K=15$, and the rank consistency of most LLM is substantially improved, reaching a more reliable conclusion.

\section{Conclusion}
\label{sec:conc}

This paper proposes a novel evaluation framework for large language models named \M. Through structurally evaluating model's capability for each test objective across multiple cognitive levels and critical concepts, \M~ achieves more comprehensive, robust and consistent evaluation for LLMs. 
Experimental results demonstrate that \M~ could effectively resist the risk of data contamination and significantly improve the rank consistency across models.
The corresponding benchmarks and leaderboard will be released, which will benefit our understanding of LLMs' capabilities.
\M~ is also broadly applicable to various applications. For instance, \M~ can function as a customizable benchmark construction framework, capable of automating evaluations for any granularity of assessment objectives, please refer to Appendix~\ref{app:custom} for details and experiments.
Furthermore, our study also sheds light on the design of future principled and trustworthy instance collection and LLM evaluation protocols.

\section*{Limitations}
Considering the balance between cost, efficiency and quality for benchmark construction, we currently use GPT-3.5 for generation in this paper, which may limit the difficulty and quality of generated instances. In the future, we will introduce more powerful LLMs (e.g., GPT-4) or incorporate human to our framework, to further improve the qualify of test instances, and release the corresponding 
 updated benchmarks.
Moreover, to facilitate the assessment of our framework, we currently select to implement \M~ based on multi-choice benchmarks.
Please also kindly note that our framework can be easily adapted to other formats of benchmark such as open-end QA and multi-turn conversation, which will be included in our future work.

\section*{Acknowledgements}
We sincerely thank all anonymous reviewers for their insightful comments and valuable suggestions. This research work is supported by the National Natural Science Foundation of China under Grants no. 62122077 and no. 62106251.

\bibliography{new_structeval,custom}

\newpage
\onecolumn

\appendix

\section{Customized Benchmark Construction based on ~\M}
\label{app:custom}

\begin{figure*}[!htp]
\centering
\subfloat[Customized benchmark construction.]{\includegraphics[width=0.45\textwidth]{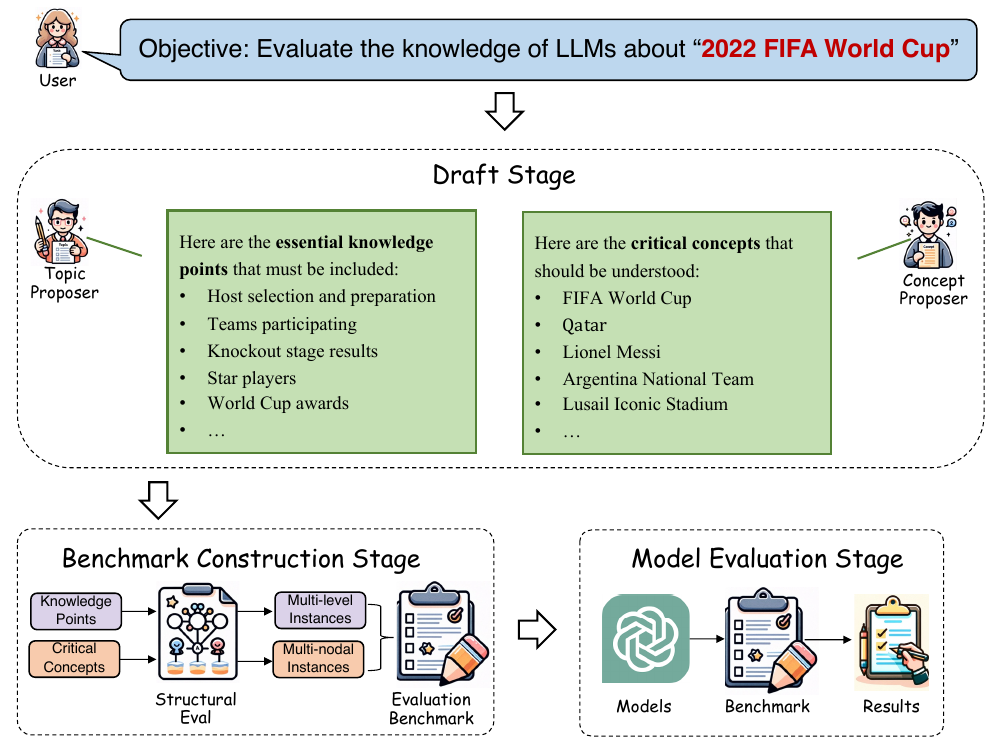}} \hspace{7pt}
\subfloat[Results on FWC\_2022.]{\includegraphics[width=0.3\textwidth]{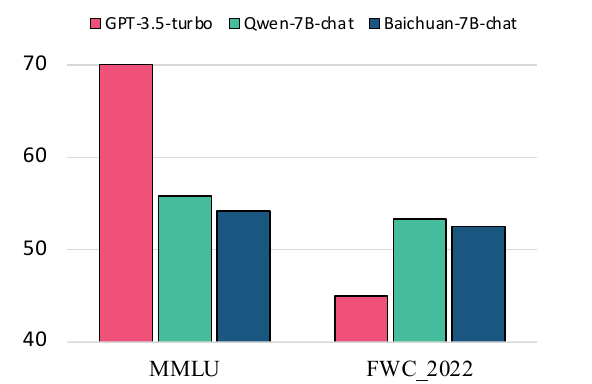}} \hspace{7pt}
\caption{In addition to expanding on existing benchmarks, \M~ can also function as a customized benchmark construction framework. It is capable of automated data construction and evaluation tailored to assessment objectives of any granularity.}
\label{fig:custom}
\end{figure*}

The majority of current benchmarks assess models in a static and coarse-grained manner. They typically start by defining a broad assessment domain, such as general knowledge, medical knowledge, or legal knowledge, and then extensively collect questions and answers within that domain. These instances are then fixedly used to evaluate models.

However, with the rapid development of large-scale models, this assessment paradigm faces two issues:
1) The obsolescence rate of static assessments is accelerating, and they are prone to rapid invalidation after reaching benchmark saturation.
2) As the application scenarios of large models become more diverse and refined, such coarse-grained assessment methods struggle to meet the rapidly growing needs for customization in real-world scenarios. For example, evaluating an AI assistant designed to aid in railway museum explanations should target "railway knowledge" rather than "general knowledge". Manually collecting data for each of these customized scenarios is not feasible.

Benefiting from the automatic and dynamic features of \M~, we can restructure it into a multi-agent-based 
 customized benchmark construction framework.
As illustrated in Figure~\ref{fig:custom}a, given a customized assessment objective (e.g., 2022 FIFA World Cup), two agents including topic proposer and concept proposer would list the essential test objectives and important concepts for comprehensive evaluate LLMs within target objective.
Then, \M~ would follow the same procedures in Figure~\ref{fig:frame}, and automatically construct a multi-level and multi-nodal benchmarks for evaluation.
In order to validate the effectiveness of our approach, we followed the aforementioned steps with GPT-4 to construct a small-scale dataset named FWC\_2022, with "2022 FIFA World Cup" as the evaluation objetive. Subsequently, we compare the performance of various models on both a large-scale general benchmark MMLU and FWC\_2022. 
FWC\_2022 comprises a total of 240 multiple-choice questions pertaining to various aspects of ``2022 FIFA World Cup.'' Please refer to the appendix for details and instances of the dataset due to page limitations.

The results in Table~\ref{fig:custom}b demonstrate the necessity for customized fine-grained evaluations: 1) In the large-scale general benchmark MMLU, GPT-3.5-turbo perform significantly better than other two LLMs, which indicate that the GPT-3.5 has a stronger general knowledge ability.
2) However, the knowledge cuttoff of GPT-3.5-turbo is September, 2021. Therefore, in FMC\_2002, the evaluate datasets about ``2022 FIFA World Cup'', GPT-3.5-turbo perform worse than other two LLMs which are newly released.
The inconsistent conclusion between these two benchmarks indicate that previous static and fine-grained evaluation could not adapt to many scenarios, and \M~ could serve as an valuable tool for a customized, dynamic and fine-grained evaluation automatically.

\section{Examples of Test Instances}

Here is an example of generated instances by~\M.

\begin{table*}[h]
    \centering
    \small
    \resizebox{0.8\textwidth}{!}{
    \begin{tabular}{p{2cm}p{6cm}p{3.5cm}}
        \toprule
        \textbf{Level} & \textbf{Instance} & \textbf{Explanation} \\ \hline
        Remembering & What is the cranial nerve number associated with the facial nerve? \newline A.CN VII \newline B.CN VI \newline C.CN V \newline D.CN VIII \newline Correct Answer: A.CN VII  & As mentioned in the provided document, the facial nerve is also known as the seventh cranial nerve, cranial nerve VII, or simply CN VII. \\ \hdashline

        Understanding & Where does the facial nerve exit the skull? \newline A.Temporal bone \newline B.Parietal bone \newline C.Occipital bone \newline D.Frontal bone \newline Correct Answer: A.Temporal bone  & As mentioned in the document, the facial nerve typically exits the skull at the stylomastoid foramen, which is located in the temporal bone. \\ \hdashline

        Applying & Which type of lesion results in a CN VII palsy, manifested as both upper and lower facial weakness on the same side of the lesion? \newline A.Lower motor neuron lesion \newline B.Upper motor neuron lesion \newline C.Trigeminal nerve lesion \newline D.Vestibulocochlear nerve lesion \newline Correct Answer: A.Temporal bone  & As per the document, lower motor neuron lesions can result in a CN VII palsy, manifested as both upper and lower facial weakness on the same side of the lesion. \\ \hdashline

        Analyzing & What type of lesion would result in both upper and lower facial weakness on the same side of the face? \newline A.Central seven lesion \newline B.Upper motor neuron lesion \newline C.Lower motor neuron lesion \newline D.Trigeminal nerve lesion \newline Correct Answer: C.Lower motor neuron lesion  & As mentioned in the provided document, lower motor neuron lesions can result in a CN VII palsy, manifested as both upper and lower facial weakness on the same side of the lesion. \\ \hdashline

        Evaluating & What is the developmental origin of the facial nerve? \newline A.First pharyngeal arch \newline B. Second pharyngeal arch \newline C.Fifth pharyngeal arch \newline D.Third pharyngeal arch \newline Correct Answer: B. Second pharyngeal arch  & According to the document, the facial nerve is developmentally derived from the second pharyngeal arch \\ \hdashline

        Creating & Propose a study to investigate the effectiveness of various taste rehabilitation techniques in patients with facial nerve damage affecting the anterior two-thirds of the tongue. \newline A.Assess the improvement in taste discrimination following acupuncture therapy. \newline B. Compare the efficacy of electronic stimulation versus traditional flavor training \newline C.Evaluate the impact of vitamin supplementation on taste recovery \newline D.Monitor changes in taste sensation after targeted facial massage \newline Correct Answer: B & Given that the facial nerve is involved in the conveyance of taste sensations from the anterior two-thirds of the tongue, comparing electronic stimulation (mimicking natural nerve impulses) with traditional flavor training (using different flavored solutions) directly addresses the rehabilitation of taste function.  \\ \hdashline

    \bottomrule
        
    \end{tabular}}
    \caption{The generated instances about the test objective ``\textit{Facial nerve}'' which is sampled from MMLU.}
    \label{tab:post_process}
\end{table*}

\section{Framework Design Details}
\label{app:detail}

This section will introduce the more details about our framework design.

\subsection{Instance Generation based on Bloom's Taxonomy}

\subsubsection{Test Objective Extraction Instruction}

We use the following instruction to identify the underlying test objective for a seed instance.

\begin{table}[h]
    \centering
    \small
    \begin{tcolorbox}

    \textbf{\# Instruction} \\
    As an expert in education and assessment, your task is to accurately identify the test objective of the seed questions I present, and provide a brief description of that test objective. I will first provide some reference examples. Please ensure that your responses follow a consistent format in line with the provided examples.  \\
    Response in the following format:\\Test Objective: <test objective of the instance>\\Description: <description of the test objective>\\
    
    \textbf{\# Example 1} \\
    Question: During the third stage of the demographic transition model, which of the following is true? \\
    A. Birth rates increase and population growth rate is less rapid. \\
    B. Birth rates decline and population growth rate is less rapid. \\
    C. Birth rates increase and population growth rate increases. \\
    D. Birth rates decrease and population growth rate increases. \\
    Correct Answer: B \\
    \textbf{Test Objective: demographic transition} \\
    Description: In demography, demographic transition is a phenomenon and theory which refers to the historical shift from high birth rates in societies with minimal technology, to low birth rates in societies with advanced technology.\\

    \textbf{\# Input} \\
    Question: <seed question> \\
    Options: <question options> \\
    Correct Answer: <answer> \\
    Test Objective:

    \end{tcolorbox}
    \caption{Prompt design for test objective extraction.}
    \label{tab:prompt_topic}
\end{table}

\subsubsection{Details of Instance Selection and Refinement}
The post-processing modules are crucial for ensuring the quality of generated instances since there exist several issue for test instances directly generated by LLMs.
Table~\ref{tab:post_process} present a case study to demonstrate how this post-processing module to filter out candidate instances with low quality.
And Table~\ref{tab:prompt_rag} shows the instruction for retrieval-augmented generation.

\begin{table}[h]
    \centering
    \begin{tcolorbox}
    \textbf{\# Instruction} \\
    Refer to the document, select the correct answer for the multiple choice questions about {subject}.\\If you can find the correct answer in the document, response with the correct choice such as ‘A/B/C/D’. \\If you cannot find the correct answer in the document, response with 'cannot answer'\\If the choices contain more than one correct option, esponse with 'cannot answer' \\Ensure your response begin with the correct choice and do not output any other content. \\
    
    \textbf{\# Input} \\
    Document: <Support document>\\
    Question: <Generated question>\\
    Choices: <Options>
    \end{tcolorbox}
    \caption{Prompt design for RAG.}
    \label{tab:prompt_rag}
\end{table}

\begin{table*}[!tp]
    \centering
    \small
    \resizebox{0.9\textwidth}{!}{
    \begin{tabular}{lp{6cm}p{3cm}}
        \toprule
        \textbf{Issue} & \textbf{Example} & \textbf{Solution} \\ \hline
        Context-dependent & \textcolor{red}{According to the provided document}, what was the impact of Comet Shoemaker–Levy 9 in July 1994? & Prompt LLM to remove \\ \hdashline
        
        Incorrect answer& What is the principle used in satellite navigation to determine exact distances? \textcolor{red}{Generated answer: Time synchronization}   & Use RAG to filter 
        \\ 
        
        Multiple answers   & What is the most important fish species produced in fish farming worldwide? A.Tuna \textcolor{red}{B.Carp} C.Mussels \textcolor{red}{D.Salmon}                &      Use RAG to filter             \\
        
        Cannot verify &  To transform the mood of 'Seasons' into a more introspective and melancholic tone, what instrumental addition would be most appropriate for the song? (\textcolor{red}{cannot verify answer correctness based on the document})            &   Use RAG to filter   \\
        \hdashline
         Too easy instance& What is the most commonly used approach for length measurement? Generated answer: rulers
         & Construct a model pool to filer \\
    \bottomrule
        
    \end{tabular}}
    \caption{Issues exist in directly generated candidate instances by LLM. And the corresponding solutions for filtering the instances with low quality. The examples are sampled from the benchmark construct process of \M.}
    \label{tab:post_process}
\end{table*}

\subsection{Instance Expansion based on Concept Mapping}

\subsubsection{Critical Concepts Extraction}
We use the following instruction to extract the critical concepts that must be understood to correctly answer the seed question.
\begin{table}[!h]
    \centering
    \small
    \begin{tcolorbox}

    \textbf{\# Instruction} \\
    As an expert in education and assessment, your task is to identify the key concepts and their related knowledge that must be understood in order to answer a given seed question. For each seed question, list all important concepts and provide a brief description for each concept. I will first provide some reference examples. Please ensure that your responses follow a consistent format in line with the provided examples. \\
    Response in the following format, each line include an concept:\\ 
    \{'name': <concept name>, 'description': <concept description>\} \\
    
    \textbf{\# Example 1} \\
    Question: During the third stage of the demographic transition model, which of the following is true? \\
    A. Birth rates increase and population growth rate is less rapid. \\
    B. Birth rates decline and population growth rate is less rapid. \\
    C. Birth rates increase and population growth rate increases. \\
    D. Birth rates decrease and population growth rate increases. \\
    Correct Answer: B \\
    \textbf{Critical Concepts:} \\
    \{'name': 'third stage of demographic transition', 'description': 'In stage three of demographic transition...'\}
    \\
    \{'name': 'birth rates', 'description': 'Birth rate is the total number of live human births per 1,000 population...'\}
    \\
    \{'name': 'population growth rate', 'description': 'Population growth is the increase in the number of people...'\}
    \\

    \textbf{\# Input} \\
    Question: <seed question> \\
    Options: <question options> \\
    Correct Answer: <answer> \\
    Critical Concepts:

    \end{tcolorbox}
    \caption{Prompt design for critical concepts extraction.}
    \label{tab:prompt_concept}
\end{table}

\subsubsection{Helpful Knowledge Triplets Selections}

We use the following instruction to select helpful knowledge triplets from all candidates.

\begin{table}[!h]
    \centering
    \small
    \begin{tcolorbox}

    \textbf{\# Instruction} \\
    Identify up to three fact triples that are most helpful to comprehend the provided question. Refer to the provided examples for valid response. If none of them is helpful, output [None]. 
    
    \textbf{\# Example 1} \\
    <Start of Question> \\
    Question: Which of the following best describes the structure that collects urine in the body? \\
    A. Bladder B. Kidney C. Ureter D. Urethra \\
    <End of Question> \\
    Candidate Triplets: \\
    1. (urine, subclass of, secretion or excretion) \\
    2. (urine, UMLS CUI, C2963137) \\
    3. (urinary bladder, connects with, urethra) \\
    4. (urinary bladder, part of, urinary system) \\
    5. (urinary bladder, subclass of, particular anatomical entity) \\
    Relevant Triplets: \\
    3. (urinary bladder, connects with, urethra) \\
    4. (urinary bladder, part of, urinary system) \\

    \textbf{\# Input} \\
    Question: <seed question> \\
    Candidate Triplets: <candidate triplets> \\
    Relevant Triplets:

    \end{tcolorbox}
    \caption{Prompt design for helpful knowledge triplets selection.}
    \label{tab:prompt_helpful}
\end{table}

\newpage

\section{Human Annotation Guidelines}

\begin{figure}[h]
    \centering
    \includegraphics[width=\textwidth]{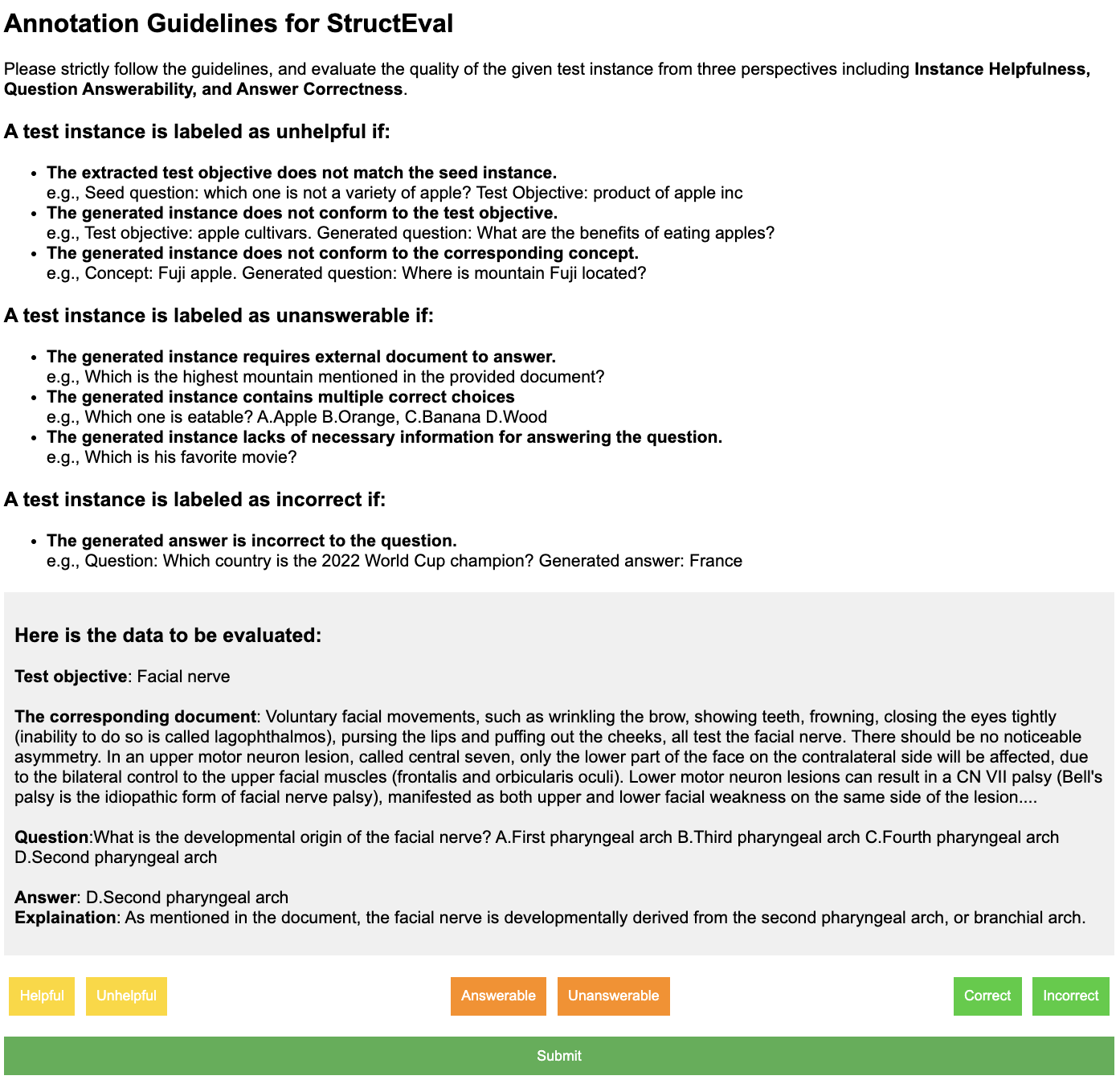}
    \caption{The annotation guidelines for our human evaluation.}
    \label{fig:guideline}
\end{figure}

Here is the annotation guidelines for our human evaluation is shown in Figure~\ref{fig:guideline}.
We recruit 5 annotators to participate in the human evaluation, each of whom possesses a bachelor degree. To ensure the clarity and consistency in the evaluation, we provided detailed instructions and examples in the annotation guidelines. Each instance is annotated by 3 participants, and the final results are determined by a majority vote.

\section{Error Analysis of Constructed Benchmark}
\label{app:error}

As we discussed in Section~\ref{sec:bench}, according to the human evaluation results, there still exist a fewer instances which not meet the standard.
In order to find the underlying causes of these errors, we conduct a detailed error analysis which is demonstrated in Table~\ref{tab:error1}, \ref{tab:error2}, and \ref{tab:error3}.

\begin{table}[!h]
    \centering
    \begin{tabular}{p{3cm}p{8cm}p{3cm}}
         \toprule
        \textbf{Test Objective} & \textbf{Example} & \textbf{Cause} \\
        \hline
        \multicolumn{3}{c}{\cellcolor[HTML]{EFEFEF}\textbf{Answerability Issue}} 
        \\
         Economic growth & Question: What constraints to economic growth \textcolor{red}{are highlighted?} A. Government intervention \quad B. Depleted resources and energy consumption \quad C. Technological advancements \quad D. Increased labor force & Context-dependent instance not be filtered \\ \hline
        Moon & What will happen in 50 billion years \textcolor{red}{according to the text?} A. The Moon's rotation will stop \quad B. The Sun will become a red giant \quad C. The Moon will collide with Earth \quad D. The Earth's rotation will match the Moon's orbital period & Context-dependent instance not be filtered \\ \hline
        Dementia & What do physicians need to include in any dementia evaluation? \textcolor{red}{A.A memory assessment  \quad B.A somatic disturbance evaluation  \quad C. A depression screening \quad D.A sensory function test} & Instance with multi-answers not be filtered by RAG \\ \hline
        Fish farming & What is the most important fish species produced in fish farming worldwide? \textcolor{red}{A.Tuna} B.Carp C.Mussels \textcolor{red}{D.Salmon} & Instance with multi-answers not be filtered by RAG \\ \hline
        Human subject research & Question: What are the three guidelines that serve as the baseline for the report? A. Beneficence, justice, and respect for persons \quad B. Beneficence, integrity, and respect \quad C. Prudence, integrity, and honesty \quad D. Beneficence, justice, and honesty & Unclear reference \\ \bottomrule
    \end{tabular}
    \caption{Error analysis of answerability issues for benchmark constructed by \M.}
    \label{tab:error1}
\end{table}

\begin{table}[!h]
    \centering
    \begin{tabular}{p{3cm}p{8cm}p{3cm}}
         \toprule
        \textbf{Test Objective} & \textbf{Example} & \textbf{Cause} \\
        \hline
                \multicolumn{3}{c}{\cellcolor[HTML]{EFEFEF}\textbf{Helpfulness Issue}} \\
        Premises Liability & \textcolor{red}{In 'peralta v. henriquez',} what was the specific dangerous condition that led to the accident? A.Lack of illumination B.Lack of security personnel C.Poor maintenance of the premises D.Inadequate signage & Unhelpful for test objective assessment \\ \hline
        psychology & Who is considered \textcolor{red}{the first director of Harvard's psychological laboratory} and a student of Wilhelm Wundt? A. Scott Lilienfeld B. Saul Kassin C. Thomas Bond D. Hugo Münsterberg & Unhelpful for test objective assessment \\ \hline
         GDP calculation & How often does \textcolor{red}{India change the base year for its GDP calculation}, according to the Frontier Strategy Group? A. Every 10 years B. Every 3 years C. Every 5 years D. Every 7 years & Unhelpful for test objective assessment \\ 
        \bottomrule
    \end{tabular}
    \caption{Error analysis of Helpfulness issues for benchmark constructed by \M.}
    \label{tab:error2}
\end{table}

\begin{table}[!h]
    \centering
    \begin{tabular}{p{3cm}p{8cm}p{3cm}}
         \toprule
        \textbf{Test Objective} & \textbf{Example} & \textbf{Cause} \\
        \hline
                \multicolumn{3}{c}{\cellcolor[HTML]{EFEFEF}\textbf{Correctness Issue}} \\

        States' rights & What was the outcome of the U.S. Supreme Court's decision in the case of California Proposition 14, and what did this decision overturn? \textcolor{red}{A. The decision overturned Proposition 14, allowing discrimination in housing.} B. The decision upheld the Rumsford Fair Housing Act, allowing discrimination in housing. C. The decision upheld Proposition 14, banning discrimination in housing. D. The decision overturned the Rumsford Fair Housing Act, banning discrimination in housing. & Conflict with document \\ \hline
        Solar System & How do the inner planets in the Solar System differ from the inferior planets? \textcolor{red}{A. The inner planets are closer to the Sun than the inferior planets.} B. The inner planets are larger in size compared to the inferior planets. C. The inner planets have atmospheres substantial enough to generate weather, while the inferior planets do not. D. The inner planets are composed mainly of gases, while the inferior planets are composed mostly of rocky materials. & Incorrect answer \\ \hline
        Physical Weathering & What type of physical weathering is considered the most important? A. Thermal fracturing \textcolor{red}{B. Frost wedging} C. Pressure release D. Wedging by plant roots & Incorrect answer 
        \\ \bottomrule
    \end{tabular}
    \caption{Error analysis of Correctness issues for benchmark constructed by \M.}
    \label{tab:error3}
\end{table}

\end{document}